%% file: main.tex
\pdfoutput=1
\documentclass[11pt]{article}

\usepackage[]{acl}
\usepackage{times}
\usepackage{latexsym}
\usepackage[T1]{fontenc}
\usepackage[utf8]{inputenc}
\usepackage{hyperref}       
\usepackage{url}            
\usepackage{booktabs}       
\usepackage{amsfonts}       
\usepackage{nicefrac}       
\usepackage{microtype}      
\usepackage{xcolor}         
\usepackage{tabularx} 
\usepackage{enumerate}
\usepackage{xspace}
\usepackage{amsthm}
\usepackage{color}
\usepackage{amsmath}
\usepackage{float}
\usepackage{algorithm}
\usepackage{algpseudocode}
\usepackage{bm}
\usepackage{multirow}
\usepackage{array}
\usepackage{makecell}
\usepackage{graphicx}
\usepackage{makecell}
\usepackage{soul}
\usepackage{bbding}
\usepackage[normalem]{ulem}
\usepackage{appendix}
\usepackage{enumitem}
\usepackage{colortbl}
\usepackage{arydshln}
\usepackage{inconsolata}
\usepackage{CJKutf8}
\usepackage[T1]{fontenc}
\usepackage[utf8]{inputenc}

\title{STATE ToxiCN: A Benchmark for Span-level Target-Aware Toxicity Extraction in Chinese Hate Speech Detection}

 \author{Zewen Bai$^{\mathbf{a}}$, Shengdi Yin$^{\mathbf{a}}$, Junyu Lu$^{\mathbf{a}}$, Jingjie Zeng$^{\mathbf{a}}$ \\
	\textbf{Haohao Zhu$^{\mathbf{a}}$, Yuanyuan Sun$^{\mathbf{a}}$, Liang Yang\thanks{* Corresponding Author}$\:\:^{\mathbf{a}}$, Hongfei Lin$^{\mathbf{a}}$} \\
	$^{\mathbf{a}}$School of Computer Science and Technology, Dalian University of Technology, China \\ 
	\texttt{dlutbzw@mail.dlut.edu.cn,}
	\texttt{(liang,hflin)@dlut.edu.cn}}

\begin{document}
\begin{CJK*}{UTF8}{gbsn}
\maketitle

\begin{abstract}
\input{Sec_Abstract.tex}
\end{abstract}

\input{Sec_Introduction.tex}
\input{Sec_RelatedWork.tex}

\input{Sec_Dataset_Construction.tex}
\input{Sec_Experiment.tex}

\input{Sec_Results.tex}
\input{Sec_Conclusion.tex}

\bibliography{main}
\bibliographystyle{acl_natbib}

\appendix

\input{appendix}

\end{CJK*}
\end{document}

%% file: Sec_Abstract.tex
The proliferation of hate speech has caused significant harm to society. The intensity and directionality of hate are closely tied to the target and argument it is associated with. However, research on hate speech detection in Chinese has lagged behind, and existing datasets lack span-level fine-grained annotations. Furthermore, the lack of research on Chinese hateful slang poses a significant challenge. In this paper, we provide two valuable fine-grained Chinese hate speech detection research resources. First, we construct a \textbf{S}pan-level \textbf{T}arget-\textbf{A}ware \textbf{T}oxicity \textbf{E}xtraction dataset (\textsc{STATE ToxiCN}), which is the first span-level Chinese hate speech dataset. Secondly, we evaluate the span-level hate speech detection performance of existing models using \textsc{STATE ToxiCN}. Finally, we conduct the first study on Chinese hateful slang and evaluate the ability of LLMs to understand hate semantics. Our work contributes valuable resources and insights to advance span-level hate speech detection in Chinese. \footnote{Code and datasets are publicly available at https://github.com/shenmeyemeifashengguo/STATE-ToxiCN.}

\textit{\textbf{Disclaimer:} The samples presented by this paper may be considered offensive or vulgar.}

%% file: Sec_Introduction.tex
\section{Introduction}


With the popularity of social media, user-generated content has experienced explosive growth, and hate speech has also flourished. Hate speech refers to harmful statements that express hatred or incite harm against specific groups or individuals based on race, religion, gender, region, sexual orientation, physical characteristics, and other factors \cite{bilewicz2020hate}. Due to its damaging impact on individuals and society, hate language is now widely considered as a problem of increasing importance \cite{silva2016analyzing}. Recently, researchers have been actively engaged in the study of hate speech detection, and this research has gradually shifted from post-level \cite{ahn2024sharedcon, alkhamissi-etal-2022-token} to span-level  \cite{pavlopoulos2021semeval, mathew2021hatexplain, zampieri2023target}.


Approximately 941 million people, or 12\% of the global population, speak Mandarin Chinese as their first language \cite{eberhard2024ethnologue}. However, research on Chinese hate speech detection lags significantly behind. There are two key issues that remain unresolved. Firstly, existing research \cite{deng2022cold, swsr, CDial} is limited to the post-level, leaving span-level Chinese hate speech detection unexplored. The intensity and directionality of hate speech are closely tied to the target and argument it is associated with \cite{cowan1996judgments}. In the Chinese linguistic context, challenges in span-level hate speech detection arise due to flexible word order and the absence of word delimiters like spaces. These issues complicate the identification of targets and arguments. For example, as shown in Exp.1 in Table \ref{intro}, the use of inversion in Chinese disrupts the conventional subject-verb-object structure. To fill this gap, we construct a span-level Chinese hate speech dataset by annotating the Target-Argument-Hateful-Group quadruples in posts.

\begin{table*}[h]
  \centering
  \resizebox{\textwidth}{!}{
    \begin{tabular}{clllll}
    \toprule
    \multicolumn{1}{l}{Exp.} & Post  & Target & Argument & Hateful & Group  \\
    \midrule
    \multirow{2}[2]{*}{1} & 你这头蠢驴，没人会喜欢。 & 你     & 蠢驴  & \multirow{2}[2]{*}{non-hate}  & \multirow{2}[2]{*}{non-hate}  \\
          & \textit{No one will ever like you, you idiot.} & \textit{you}   & \textit{idiot} &       &  \\
    \midrule
    \multirow{2}[2]{*}{2} & 男同是艾滋高发群体。 & 男同    & 艾滋高发群体  & \multirow{2}[2]{*}{hate} & \multirow{2}[2]{*}{LGBTQ, others}  \\
          & \textit{Gay people are a high-risk group for HIV.} & \textit{Gay people} & \textit{a high-risk group for HIV} &       &  \\
    \midrule
    \multirow{2}[2]{*}{3} & 默我是真的很讨厌。 & 默(黑犬) & 讨厌   
 & \multirow{2}[2]{*}{hate} & \multirow{2}[2]{*}{Racism}  \\
          & \textit{Silence, I really hate it.} & \textit{Slience(black dog)} & \textit{hate}  &       &  \\
    \bottomrule
    \end{tabular}}
    \caption{Examples of annotated posts from \textsc{STATE ToxiCN} dataset with corresponding annotations of Target-Argument-Hateful-Group quadruples.}
  \label{intro}%
\end{table*}

Secondly, although Chinese hate lexicons have been developed \cite{swsr, toxicn}, the absence of interpretable annotations hinders a deeper understanding of hate speech. As an ideographic language, Chinese has a wealth of synonyms and near-synonyms \cite{mair1991chinese}, making the forms and types of hateful slang diverse and difficult to capture. Chinese hateful slang often evades model detection through techniques such as homophonic substitution, character splitting and merging, and historical allusions \cite{toxicloakcn}. For example, in Exp. 3 from Table \ref{intro}, the target uses the merging technique to evade detection. To address this issue, we collect commonly used hateful slang from real-world online forums, providing detailed annotations to construct the first annotated lexicon for Chinese hateful slang. This lexicon serves as a foundational resource for research into the understanding of Chinese hate semantics.

To address the lack of resources for span-level Chinese hate speech research, we introduce a \textbf{S}pan-level \textbf{T}arget-\textbf{A}ware \textbf{T}oxicity \textbf{E}xtraction dataset (\textsc{STATE ToxiCN}), a novel dataset containing 8029 posts and 9533 Target-Argument-Hateful-Group quadruples addressing sexism, racism, regional bias, and anti-LGBTQ sentiments. Using this dataset, we evaluate the performance of LLMs in span-level Chinese hate speech detection. Specially, we annotate: 1) extraction of the targets and arguments from the post, 2) determination of whether each Target-Argument pair constitutes hate speech, and 3) classification of the groups for hateful Target-Argument pairs. 

Moreover, we summarize and annotate commonly used hateful slang from real-world online forums to address the challenges of identifying hateful slang. We compile a comprehensive annotated lexicon, labeling each hateful slang with its frequent groups and providing explanations of its usage and contextual meaning. This is the first annotated lexicon of Chinese hateful slang with interpretative annotations. This resource not only helps to understand how hateful slang is used to disguise hate speech, but also provides valuable annotated data to evaluate and improve the ability of LLMs to detect hateful slang. The main contributions of this work are summarized as follows:

\begin{itemize}
\setlength\itemsep{0em}

\item We provide a span-level target-aware toxicity extraction dataset, containing 8k posts and 9.5k quadruples, filling the gap in span-level resources for Chinese hate speech detection.

\item We construct an annotated lexicon of commonly used hateful slang in real-world online forums to evaluate LLMs' capability in understanding Chinese hate speech semantics.

\item We evaluate models on \textsc{STATE ToxiCN}, assessing their span-level performance and ability to detect hateful slang, highlighting key challenges and insights for improvement.

\end{itemize}

%% file: Sec_RelatedWork.tex
\section{Related Work}

\begin{table*}[t]
	\small
	\centering
	\renewcommand{\arraystretch}{1.1} 
	\setlength{\tabcolsep}{5pt} 
	\resizebox{\textwidth}{!}{
	\begin{tabular}{lllrccccc}
		\toprule
		\textbf{Work} & \textbf{Platforms} & \textbf{Language} & \textbf{\#Posts} & \textbf{Span} & \textbf{Tar.} & \textbf{Arg.} & \textbf{Group} & \textbf{Lex.} \\
		\midrule
		\citet{davidson2017automated} & Twitter & English & 24,802 &  &  &  &  & \checkmark \\
		\citet{founta2018large} & Twitter & English & 80,000 &  &  &  &  &  \\
		Toxic Spans \cite{pavlopoulos2021semeval} & Civil Comments & English & 10,629 & \checkmark &  &  &  &  \\
		HateXplain \cite{mathew2021hatexplain} & Twitter, Gab & English & 20,148 & \checkmark &  &  & \checkmark &  \\
		TBO \cite{zampieri2023target} & Twitter & English & 4,673 & \checkmark & \checkmark & \checkmark &  &  \\
		\hdashline
		COLD \cite{deng2022cold} & Zhihu, Weibo & Chinese & 37,480 &  &  &  & \checkmark &  \\
		SWSR \cite{swsr} & Weibo & Chinese & 8,969 &  &  &  & \checkmark & \checkmark \\
		Cdial-Bias-Utt \cite{CDial} & Zhihu & Chinese & 13,394 &  &  &  & \checkmark &  \\
		Cdial-Bias-Ctx \cite{CDial} & Zhihu & Chinese & 15,013 &  &  &  & \checkmark &  \\
		\textsc{ToxiCN} \cite{toxicn} & Zhihu, Tieba & Chinese & 12,011 &  &  &  & \checkmark & \checkmark \\
		\textsc{ToxiCloakCN} \cite{toxicloakcn} & Zhihu, Tieba & Chinese & 4,582 &  &  &  & \checkmark &  \\
		\hdashline
		\textsc{STATE ToxiCN} (Ours) & Zhihu, Tieba & Chinese & 8,029 & \checkmark & \checkmark & \checkmark & \checkmark & \checkmark \\
		\bottomrule
	\end{tabular}}
	\vspace{-0.2cm}
	\caption{Comparison of hate speech datasets based on \textit{Platforms}, \textit{Language}, \textit{\#Posts}, span-level annotations (\textit{Span}), inclusion of Target (\textit{Tar.}), Argument (\textit{Arg.}), \textit{Group}, and Lexicon information (\textit{Lex.}).}
	\label{dataset_work}%
\end{table*}

\paragraph{Hate Speech Detection.}
Hate speech detection is a critical task in Natural Language Processing (NLP) that has attracted considerable attention recently. Researchers have increasingly turned to pre-trained models to address this issue \cite{caselli2020hatebert, Detoxify, zhou2021hate, alkhamissi2022token, ali2022hate}. To facilitate progress in this field, several datasets tailored to hate speech detection have been developed \cite{waseem2016hateful, davidson2017automated, founta2018large, hartvigsen2022toxigen}. \citet{pavlopoulos2021semeval} introduced span-level hate speech detection, while the TBO dataset advanced the field by pioneering the extraction of Target-Argument-Harmful triples \cite{zampieri2023target}. However, Chinese hate speech detection remains significantly underdeveloped.

\paragraph{Chinese Hate Speech Dataset.}
While some Chinese hate speech datasets exist, these efforts remain limited to the post-level. TOCP and TOCAB, derived from Taiwan's PTT platform, focus on detecting profanity and abusive language \cite{tocab}. The Sina Weibo Sexism Review (SWSR) centers on identifying sexism \cite{swsr}. The Chinese Offensive Language Dataset (COLD) categorizes sentences into types such as individual attacks and anti-bias \cite{deng2022cold}. \citet{CDial} introduce CDial-Bias, the first annotated dataset specifically addressing social bias in Chinese dialogues. \citet{toxicn} present \textsc{ToxiCN}, a dataset encompassing both explicit and implicit toxic language samples. \citet{toxicloakcn} introduce a dataset to evaluate LLMs' robustness against cloaking perturbations.

While previous works offer quality corpora, span-level research remains unexplored. \textsc{STATE ToxiCN} is the first such dataset, annotating Target-Argument-Hateful- Group quadruples. Unlike existing lexicons, we offer interpretative annotations and targeted group labels, creating a comprehensive Chinese hateful slang lexicon.

%% file: Sec_Dataset_Construction.tex
\section{Dataset Construction}

\subsection{Overview}

In this section, we introduce the construction process of the \textsc{STATE ToxiCN} dataset and the annotated lexicon of Chinese hateful slang. First, we describe the data sources and filtering procedures. Next, we detail the annotation process and the measures implemented to ensure high annotation quality. We then examine the Inter-Annotator Agreement (IAA) at different levels of granularity. Finally, we present relevant statistics for the \textsc{STATE ToxiCN} dataset.

\subsection{Data Source and Filtering}

Our dataset construction is based on the post-level Chinese hate speech dataset \textsc{ToxiCN} \cite{toxicn}. We develop a high-quality span-level Chinese hate speech dataset through data filtering and annotation processes. During the data filtering phase, we extract potential samples from the original dataset and systematically clean low-quality, incomplete, or irrelevant content. For instance, we remove meaningless text such as advertisements, random character combinations, and overly short (less than 5 characters) or overly long (more than 500 characters) text fragments.

Building on this, we conduct a comprehensive review of the hate speech annotations, particularly regarding the clarity of the target and argument components of hate speech. For example, in cases describing strongly discriminatory or biased content, we further examine whether the target was specific, removing ambiguous or undefined samples. In particular, for texts lacking a clear Target-Argument structure, we opt to exclude them to ensure the dataset could accommodate the requirements of span-level analysis.

\subsection{Data Annotation}
\subsubsection{Annotation Guidelines}

During the annotation process, we establish guidelines and implement multi-stage quality control to ensure the consistency of the annotations, aiming to label Target-Argument-Hateful-Group quadruples in the posts. First, we develop detailed annotation guidelines, including standards for extracting targets and arguments (Target-Argument Pair), criteria for determining hatefulness (Hateful), and methods for classifying groups (Group):

\begin{description}[leftmargin=0.5cm]
	\setlength{\itemsep}{0.2em}
	
	\item[\textbf{Target-Argument Pair}] A span consists of both the target and its corresponding argument extracted from the post. A single post may contain more than one Target-Argument Pair.
	
	\item[\textbf{Hateful}]   If the Target-Argument Pair explicitly or implicitly conveys harm towards the target or other groups, it is labeled as "\textbf{hateful}"; otherwise, it is labeled as "\textbf{non-hate}."
	
	\item[\textbf{Group}] Building on the Target-Argument Pair, this category identifies specific groups targeted by hateful expressions, with a single pair probably involving multiple groups.

\end{description}

For example, in Exp. 2 from Table \ref{intro}, Target-Argument Pair is ``男同（\textit{Gay people}）'' and ``艾滋高发群体(\textit{a high-risk group for HIV})''. This Target-Argument Pair constitutes hate against the LGBTQ community and people living with AIDS, so the Hateful label is ``\textit{hate}'' and the Group label is ``\textit{LGBTQ, others}''. The quadruple is [\textit{ 男同 | 艾滋高发群体 | hate | LGBTQ, others }].

Regarding the annotation of the Chinese hateful slang lexicon, we also establish annotation guidelines for identifying, categorizing, and labeling the terms, with a focus on their frequent groups (Group) and contextual explanations (Explanation):

\begin{description}[leftmargin=0.5cm]
	\setlength{\itemsep}{0.2em}
	
	\item[\textbf{Group}] Each hateful slang term is categorized by the group it targets, such as sexism, racism, regional bias, anti-LGBTQ, or others. Some terms may target multiple groups.

	\item[\textbf{Explanation}] An explanation of hateful slang is provided, including its literal meaning, extended meanings, the reasons for hatred towards targeted groups, and common usage patterns.
\end{description}

\subsubsection{Mitigating Bias.} 

To mitigate bias, we assemble annotators with diverse backgrounds, including differences in gender, age, ethnicity, region, and educational level. The detailed information of annotators can be found in Appendix \ref{ann}. All annotators possess linguistic expertise and have undergone systematic training. During the annotation process, we primarily employ regular cross-validation and expert arbitration to ensure the objectivity and consistency of annotation results. Additionally, we maintain an online document to record and update Chinese hateful slang identified in the posts. This annotated lexicon serves not only as a research resource but also helps to align annotators' standards.

\subsubsection{Annotation Procedure}

\begin{table}[t]
    \centering
    \resizebox{\columnwidth}{!}{%
        \begin{tabular}{cccc} 
            \toprule
            \textbf{Target Span} & \textbf{Argument Span} & \textbf{If Hateful} & \textbf{Targeted Group} \\
            \midrule
            0.65 & 0.61 & 0.68 & 0.75 \\
            \bottomrule
        \end{tabular}
    }
    \caption{Fleiss’ Kappa for different granularities.}
    \label{kappa}
    \vspace{-0.3cm}
\end{table}

\paragraph{Cross-validation and Expert Arbitration.} Each text is independently annotated by at least two annotators, who followed a unified annotation guideline to extract Target-Argument pairs, determine the hatefulness, and classify the groups. After the initial annotation phase, 20\% of the samples are regularly selected for cross-validation, during which other annotators re-annotate these samples to ensure a consistent understanding of the rules and standards across annotators. This approach allows us to identify and resolve potential biases or discrepancies in the annotations in a timely manner. 

For disputed samples with significant annotation differences, an arbitration team of domain experts reviews the samples, considering the textual context and annotation guidelines to determine the most appropriate annotation. We explore inter-annotator agreement (IAA) on \textsc{STATE ToxiCN}, with Fleiss' kappa scores \cite{fleiss1971measuring} for each hierarchy detailed in Table \ref{kappa}. The detailed analysis can be found in Appendix \ref{app_kappa}.

\begin{table}[t]
	\centering
	\resizebox{\columnwidth}{!}{
		\begin{tabular}{llcc}
			\toprule
			\textbf{Category} & \textbf{Subcategory} & \textbf{Count} & \textbf{Percentage (\%)} \\
			\midrule
			\multirow{5}{*}{Groups} & Gender       & 1663 & 17.44 \\
			& Race         & 1232 & 12.92 \\
			& Region       & 1323 & 13.88 \\
			& LGBTQ        & 628  & 6.59  \\
			& Others       & 351  & 3.68  \\
			& Multi-group   & 866  & 9.08  \\
			\midrule
			\multirow{2}{*}{Hateful}     & Hate         & 6063 & 63.60 \\
			& Non-Hate     & 3470 & 36.40 \\
			\midrule
			\textbf{Total}                  & -            & 9533 & 100.00 \\
			\bottomrule
		\end{tabular}}
	\caption{Statistics of annotated posts from the \textsc{STATE ToxiCN} dataset, including Target, Argument, Group, and Hatefulness classifications.}
	\label{datasetlabel}
    \vspace{-0.7cm}
\end{table}

\paragraph{Lexicon Annotation.} To ensure annotation quality, we maintain a shared online document for recording and updating detailed information on Chinese hateful slang. This document includes explanations of each slang term and the specific groups they typically reference. Team members could add newly discovered slang terms, which are then evaluated and annotated by the expert team. This dynamic maintenance ensures a consistent understanding of emerging language and slang among all team members.

Additionally, the shared online document serves as a knowledge-sharing platform, providing the annotation team with consistent references and concrete examples for handling complex or ambiguous posts. This mechanism improves annotation efficiency and enhances the consistency of the annotations. This is the first Chinese hateful slang lexicon with interpretable annotations. This lexicon provides valuable research resources for span-level Chinese hate speech semantic understanding.

\subsection{Data Description}

\textsc{STATE ToxiCN} dataset contains a total of 8,029 annotated posts, among which 4,942 posts include hateful content, accounting for 61.55\%. A total of 9,533 quadruples are annotated, with 6,034 of them containing hateful information, making up 63.60\%. We present the statistical details of \textsc{STATE ToxiCN} in Table \ref{datasetlabel}. Gender, Region, and Race are the three most common group types in the dataset. Additionally, "multi-group" refers to Target-Argument pairs that involve hatred directed at multiple target groups, with a total of 854 such instances, accounting for 8.96\%. In addition, the annotated lexicon includes 830 Chinese hateful slang terms collected from real online forums. More annotated examples can be found in Appendix \ref{lexicon}.

%% file: Sec_Experiment.tex
\section{Experiment}

\subsection{Baselines}
To evaluate the performance of LLMs with varying parameter sizes and capabilities in detecting span-level Chinese hate speech, we choose twelve well-known models across three categories:

\noindent \textbf{Open-source Models:} mT5-base \cite{mt5}, Mistral-7B \cite{mistral}, LLaMA3-8B \cite{llama3modelcard}, Qwen2.5-7B \cite{qwen2.5}; \textbf{Safety-domain Models:} ShieldLM-14B-Qwen \cite{shieldlm}, ShieldGemma-9B \cite{shieldgemmagenerativeaicontent}, and \textbf{Closed-source LLMs:} LLaMA3-70B \cite{llama3modelcard}, Qwen2.5-72B \cite{qwen2.5}, Gemini-1.5-Pro \cite{gemini}, Claude-3.5-Sonnet \cite{claude3.5}, GPT-4o \cite{gpt4o}, DeepSeek-v3 \cite{liu2024deepseek}. Detailed information of fine-tuning is provided in Appendix \ref{fin}.

\subsection{Evaluation metrics.}

Due to the ambiguity of Chinese span boundaries, a single evaluation metric may not accurately assess the performance of models in span-level Chinese hate speech detection. To obtain more accurate evaluation results, we therefore utilize both hard and soft matching metrics. We employ the Macro-F1 scores as the main evaluation metrics. 

\noindent \textbf{Hard-matching:} A predicted quadruplet, particularly its target and argument components, is deemed correct only if it perfectly matches its corresponding gold label.

\noindent \textbf{Soft-matching:} We adopt the algorithm proposed by \citet{soft}, where a prediction is considered correct if the Target and Argument scores achieve a threshold of 0.5.

\begin{table}[ht]
	\centering
	\resizebox{\columnwidth}{!}{
		\begin{tabular}{lcccc}
			\toprule
			\textbf{Category} & \textbf{\#Posts} & \textbf{Quad.} & \textbf{Hateful} & \textbf{Non-hate} \\
			\midrule
			Train & 6424 & 7631 & 4842 & 2789 \\
			Test & 1605  & 1902 & 1221 & 681 \\
			\midrule
			\textbf{Total}     & 8029 & 9533 & 6063 & 3470 \\
			\bottomrule
	\end{tabular}}
	\caption{Statistics of train and test datasets, including \textit{\#Posts}, total quads (\textit{Quad.}), hateful quads (\textit{Hateful}), and non-hateful quads (\textit{Non-hate}).}
	\label{train}
    \vspace{-0.5cm}
\end{table}

\subsection{Experiment Settings}

We conduct fine-tuning on open-source and safety-domain models, with training and testing set sizes detailed in Table \ref{train}. The fine-tuning process was performed using only a basic prompt, which included task definitions, output formats, and specific prediction requirements for all elements. We evaluate the performance of closed-source LLMs by calling their APIs. In addition to the aforementioned basic prompt, we also provided a hate speech example and a non-hate speech example. Further details regarding these requirements are provided in Appendix \ref{prompt}.

\definecolor{intnull}{RGB}{232,228,241}
\definecolor{intnu}{RGB}{225,233,246}

%% file: Sec_Results.tex
\section{Results and Analysis}

\definecolor{intnull}{RGB}{232,228,241}
\definecolor{intnu}{RGB}{225,233,246}

\begin{table*}[t]
	\centering
	\small
	\renewcommand{\arraystretch}{1.05} 
	\setlength{\tabcolsep}{5pt} 
	\resizebox{\textwidth}{!}{
		\begin{tabular}{lcccccccccc}
			\toprule
			\multirow{2}{*}{\textbf{Model}} 
			& \multicolumn{2}{c}{\textbf{Target}} 
			& \multicolumn{2}{c}{\textbf{Argument}} 
			& \multicolumn{2}{c}{\textbf{T-A Pair}} 
			& \multicolumn{2}{c}{\textbf{T-A-H Tri.}} 
			& \multicolumn{2}{c}{\textbf{Quad.}} \\
			\cmidrule(lr){2-3} \cmidrule(lr){4-5} \cmidrule(lr){6-7} \cmidrule(lr){8-9} \cmidrule(lr){10-11}
			& \textbf{Hard} & \textbf{Soft} & \textbf{Hard} & \textbf{Soft} 
			& \textbf{Hard} & \textbf{Soft} & \textbf{Hard} & \textbf{Soft} 
			& \textbf{Hard} & \textbf{Soft} \\
    		\midrule
			\rowcolor{white} \multicolumn{11}{c}{\textit{Finetuned Models (with Basic Prompt)}} \\
			\midrule
			mT5-base   & 59.15 & 70.55 & 28.63 & 67.03 & 23.33 & 55.90 & 17.76 & 43.34 & 16.60 & 38.61 \\
			Mistral-7B    & 62.97 & 73.69 & \underline{35.58} & \underline{70.90} & 30.55 & 60.49 & 26.15 & 51.01 & \underline{23.72} & 45.62 \\
			LLaMA3-8B     & \textbf{64.07} & 73.74 & \textbf{36.72} & 70.82 & \textbf{31.64} & \underline{60.88} & \textbf{27.04} & 51.62 & \textbf{24.27} & 46.08 \\
			Qwen2.5-7B     & \underline{63.96} & \textbf{74.64} & 35.42 & 70.36 & \underline{30.63} & 60.52 & \underline{26.51} & \textbf{52.86} & 23.70 & \underline{47.03} \\
			\hdashline
			ShieldLM-14B-Qwen       & 63.83 & 73.45 & 34.80 & 70.23 & 30.20 & 59.81 & 26.18 & 51.24 & 23.59 & 45.58 \\  	
			ShieldGemma-9B       & 63.40 & \underline{74.31} & 34.40 & \textbf{71.11} & 29.99 & \textbf{61.51} & 25.64 & \underline{52.70} & 23.49 & \textbf{47.14} \\ 	
    		\midrule
			\rowcolor{white} \multicolumn{11}{c}{\textit{LLM APIs (with Basic Prompt and 2 Examples)}} \\
			\midrule
			LLaMA3-70B     & 30.54 & 41.03 & 14.39 & 47.96 & 8.16 & 27.34 & 6.03 & 20.70 & 3.69 & 11.93 \\
			Qwen2.5-72B     & 40.94 & 50.44 & 21.10 & 56.36 & 15.66 & 39.49 & 12.48 & 30.92 & 8.74 & 20.29 \\
			Gemini-1.5-Pro      & 29.80 & 37.29 & 18.43 & 54.96 & 9.37 & 26.22 & 7.71 & 21.88 & 5.45 & 14.81 \\
			Claude-3.5-Sonnet   & 37.61 & 50.72 & 15.45 & 57.24 & 9.72 & 36.16 & 7.94 & 29.82 & 6.29 & 22.45 \\
			GPT-4o          & 46.85 & 58.19 & 22.64 & 62.41 & 17.21 & 46.41 & 13.21 & 35.68 & 9.00 & 23.34 \\
			DeepSeek-v3     & 48.16 & 59.25 & 22.79 & 59.38 & 18.68 & 46.40 & 14.95 & 37.19 & 11.48 & 27.38 \\
			\bottomrule
	\end{tabular}}
	\caption{Performance comparison of various models across different levels of annotated tasks, including \textit{Target}, \textit{Argument}, Target-Argument Pair (\textit{T-A Pair}), Target-Argument-Hateful Triple (\textit{T-A-H Tri.}), and Target-Argument-Hateful-Group Quadruple (\textit{Quad.}) under both Hard and Soft evaluation metrics.}
	\label{tab:comparison}
    \vspace{-0.5cm}
\end{table*}

\subsection{\textit{RQ1: Can LLMs identify the spans of targets and arguments in Chinese text?}}








\paragraph{Finetuned Models}
In the tasks of identifying targets, arguments, and target-argument pairs (\textit{T-A Pair}), fine-tuned models significantly outperform direct usage of LLM APIs. LLaMA3-8B, Qwen2.5-7B, and the Shield series models all achieve hard match scores of over 63\%, strongly demonstrating the superior ability of fine-tuned models in identifying text span boundaries. Soft match metrics further confirm the advantage of fine-tuned models in target and argument identification, with F1 score approaching 70\% or higher. These results indicate that fine-tuned models on task-specific data show a substantial advantage in span identification. LLaMA3-8B and Qwen2.5-7B perform the best. ShieldGemma-9B achieves the highest scores in soft-match metrics for two tasks.

\paragraph{LLM APIs}
Compared to fine-tuned models, LLMs access through APIs performed significantly worse in all tasks, regardless of whether hard or soft match metrics were used. Even with few-shot prompting strategies, their performance still lag far behind that of fine-tuned models. This suggests that LLMs, without being fine-tuned on Chinese hate speech data, are not adept at identifying text span boundaries, which is consistent with the fact that their training objectives are not directly related to this task. Specifically, the hard match scores for LLMs on "Target" and "Argument" were only between 40-50\%, while soft match scores were around 60\%. GPT-4o and DeepSeek-v3 exhibit the best performance.

\paragraph{Comparison and Summary}
The fine-tuning technology can improve models’ abilities in identifying text span boundaries, demonstrating a clear advantage in target and argument extraction tasks. Joint extraction tasks are more challenging than single element extraction tasks, and all models perform worse in argument extraction than in target extraction. We believe this is primarily because in Chinese text, argument spans tend to be longer and structurally more complex, leading to poor hard match performance. However, the soft match scores for argument extraction were similar to those of target extraction, indicating that the models' semantic understanding capabilities are comparable in both tasks. This suggests that despite difficulties in identifying precise boundaries, the models remain effective in semantic understanding.

\definecolor{dec}{RGB}{56,87,35}
\definecolor{rise}{RGB}{192,0,0}
\definecolor{human}{RGB}{47,85,151}

\begin{table*}[t]
	\centering
	\small
	\renewcommand{\arraystretch}{1.05} 
	\setlength{\tabcolsep}{5pt} 
	\resizebox{\textwidth}{!}{
		\begin{tabular}{lcccccccccc}
			\toprule
			\multirow{2}{*}{\textbf{Model}} 
			& \multicolumn{2}{c}{\textbf{Target}} 
			& \multicolumn{2}{c}{\textbf{Argument}} 
			& \multicolumn{2}{c}{\textbf{T-A Pair}} 
			& \multicolumn{2}{c}{\textbf{T-A-H Tri.}} 
			& \multicolumn{2}{c}{\textbf{Quad.}} \\
			\cmidrule(lr){2-3} \cmidrule(lr){4-5} \cmidrule(lr){6-7} \cmidrule(lr){8-9} \cmidrule(lr){10-11}
			& \textbf{Hard} & \textbf{Soft} & \textbf{Hard} & \textbf{Soft} 
			& \textbf{Hard} & \textbf{Soft} & \textbf{Hard} & \textbf{Soft} 
			& \textbf{Hard} & \textbf{Soft} \\
    		\midrule
			\rowcolor{white} \multicolumn{11}{c}{\textit{Finetuned Models (with Basic Prompt)}} \\
			\midrule
			mT5-base   & 56.83$_{\textcolor{dec}{2.32}}$ & 68.33$_{\textcolor{dec}{2.22}}$ & 27.17$_{\textcolor{dec}{1.46}}$ & 64.17$_{\textcolor{dec}{2.86}}$ & 21.33$_{\textcolor{dec}{2.00}}$ & 51.17$_{\textcolor{dec}{4.73}}$ & 18.17$_{\textcolor{rise}{1.46}}$ & 44.67$_{\textcolor{rise}{1.33}}$ & 16.33$_{\textcolor{dec}{0.27}}$ & 36.17$_{\textcolor{dec}{2.44}}$ \\
            Mistral-7B   & 61.03$_{\textcolor{dec}{1.94}}$ & 72.62$_{\textcolor{dec}{1.07}}$ & 36.71$_{\textcolor{rise}{1.13}}$ & 70.05$_{\textcolor{dec}{0.85}}$ & 30.27$_{\textcolor{dec}{0.28}}$ & 58.62$_{\textcolor{dec}{1.87}}$ & 26.25$_{\textcolor{rise}{0.10}}$ & 51.05$_{\textcolor{rise}{0.04}}$ & 22.22$_{\textcolor{dec}{1.50}}$ & 43.00$_{\textcolor{dec}{2.62}}$ \\
			LLaMA3-8B    & 61.26$_{\textcolor{dec}{2.81}}$ & 72.12$_{\textcolor{dec}{1.62}}$ & 35.17$_{\textcolor{dec}{1.55}}$ & 70.83$_{\textcolor{rise}{0.01}}$ & 28.53$_{\textcolor{dec}{3.11}}$ & 59.00$_{\textcolor{dec}{1.88}}$ & 24.47$_{\textcolor{dec}{2.57}}$ & 51.38$_{\textcolor{dec}{0.24}}$ & 19.77$_{\textcolor{dec}{4.50}}$ & 42.46$_{\textcolor{dec}{3.62}}$ \\
Qwen2.5-7B   & 61.79$_{\textcolor{dec}{2.17}}$ & 73.33$_{\textcolor{dec}{1.31}}$ & 36.26$_{\textcolor{rise}{0.82}}$ & 69.59$_{\textcolor{dec}{0.77}}$ & 29.92$_{\textcolor{dec}{0.71}}$ & 57.72$_{\textcolor{dec}{2.80}}$ & 27.64$_{\textcolor{rise}{1.13}}$ & 53.66$_{\textcolor{rise}{0.80}}$ & 22.76$_{\textcolor{dec}{0.94}}$ & 45.04$_{\textcolor{dec}{1.99}}$ \\
\hdashline
ShieldLM-14B-Qwen  & 64.08$_{\textcolor{rise}{0.25}}$ & 74.48$_{\textcolor{rise}{1.03}}$ & 34.19$_{\textcolor{dec}{0.61}}$ & 69.86$_{\textcolor{dec}{0.37}}$ & 28.90$_{\textcolor{dec}{1.30}}$ & 58.30$_{\textcolor{dec}{1.51}}$ & 25.60$_{\textcolor{dec}{0.58}}$ & 51.69$_{\textcolor{rise}{0.45}}$ & 21.30$_{\textcolor{dec}{2.29}}$ & 43.60$_{\textcolor{dec}{1.98}}$ \\
ShieldGemma-9B   & 62.50$_{\textcolor{dec}{0.90}}$ & 74.35$_{\textcolor{rise}{0.04}}$ & 35.71$_{\textcolor{rise}{1.31}}$ & 71.10$_{\textcolor{dec}{0.01}}$ & 29.87$_{\textcolor{dec}{0.12}}$ & 60.71$_{\textcolor{dec}{0.80}}$ & 26.95$_{\textcolor{rise}{1.31}}$ & 55.03$_{\textcolor{rise}{2.33}}$ & 23.21$_{\textcolor{dec}{0.28}}$ & 46.10$_{\textcolor{dec}{1.04}}$ \\
    		\midrule
			\rowcolor{white} \multicolumn{11}{c}{\textit{LLM APIs (with Basic Prompt and 2 Examples)}} \\
			\midrule
LLaMA3-70B    & 30.87$_{\textcolor{rise}{0.33}}$ & 41.45$_{\textcolor{rise}{0.42}}$ & 14.80$_{\textcolor{rise}{0.41}}$ & 46.68$_{\textcolor{dec}{1.28}}$ & 8.29$_{\textcolor{rise}{0.13}}$ & 25.38$_{\textcolor{dec}{1.96}}$ & 7.40$_{\textcolor{rise}{1.37}}$ & 22.58$_{\textcolor{rise}{1.88}}$ & 4.72$_{\textcolor{rise}{1.03}}$ & 13.14$_{\textcolor{rise}{1.21}}$ \\
Qwen2.5-72B   & 45.58$_{\textcolor{rise}{4.64}}$ & 54.67$_{\textcolor{rise}{4.23}}$ & 22.15$_{\textcolor{rise}{1.05}}$ & 57.36$_{\textcolor{rise}{1.00}}$ & 16.77$_{\textcolor{rise}{1.11}}$ & 41.61$_{\textcolor{rise}{2.12}}$ & 12.42$_{\textcolor{dec}{0.06}}$ & 30.35$_{\textcolor{dec}{0.57}}$ & 8.83$_{\textcolor{rise}{0.09}}$ & 19.59$_{\textcolor{dec}{0.70}}$ \\
Gemini-1.5-Pro   & 31.52$_{\textcolor{rise}{1.72}}$ & 39.07$_{\textcolor{rise}{1.78}}$ & 18.94$_{\textcolor{rise}{0.51}}$ & 54.57$_{\textcolor{dec}{0.39}}$ & 9.01$_{\textcolor{dec}{0.36}}$ & 27.95$_{\textcolor{rise}{1.73}}$ & 8.61$_{\textcolor{rise}{0.90}}$ & 25.83$_{\textcolor{rise}{3.95}}$ & 6.23$_{\textcolor{rise}{0.78}}$ & 17.35$_{\textcolor{rise}{2.54}}$ \\
Claude-3.5-Sonnet   & 41.45$_{\textcolor{rise}{3.84}}$ & 54.06$_{\textcolor{rise}{3.34}}$ & 15.80$_{\textcolor{rise}{0.35}}$ & 55.80$_{\textcolor{dec}{1.44}}$ & 10.43$_{\textcolor{rise}{0.71}}$ & 36.96$_{\textcolor{rise}{0.80}}$ & 9.28$_{\textcolor{rise}{1.34}}$ & 33.04$_{\textcolor{rise}{3.22}}$ & 7.10$_{\textcolor{rise}{0.81}}$ & 25.22$_{\textcolor{rise}{2.77}}$ \\
GPT-4o    & 49.28$_{\textcolor{rise}{2.43}}$ & 61.30$_{\textcolor{rise}{3.11}}$ & 21.71$_{\textcolor{dec}{0.93}}$ & 59.66$_{\textcolor{dec}{2.75}}$ & 16.52$_{\textcolor{dec}{0.69}}$ & 46.55$_{\textcolor{rise}{0.14}}$ & 12.01$_{\textcolor{dec}{1.20}}$ & 33.72$_{\textcolor{dec}{1.96}}$ & 8.46$_{\textcolor{dec}{0.54}}$ & 22.80$_{\textcolor{dec}{0.54}}$ \\
DeepSeek-v3    & 52.69$_{\textcolor{rise}{4.53}}$ & 64.25$_{\textcolor{rise}{4.00}}$ & 23.79$_{\textcolor{rise}{1.00}}$ & 62.10$_{\textcolor{rise}{2.72}}$ & 19.22$_{\textcolor{rise}{0.54}}$ & 50.40$_{\textcolor{rise}{4.00}}$ & 16.13$_{\textcolor{rise}{1.18}}$ & 42.07$_{\textcolor{rise}{4.88}}$ & 11.69$_{\textcolor{rise}{0.21}}$ & 30.11$_{\textcolor{rise}{2.73}}$ \\
			\bottomrule
	\end{tabular}}
	\caption{Performance comparison of various models on a hateful-slang-containing subset of the test set. Green indicates a \textcolor{dec}{decrease} in F1 score on the subset compared to the full dataset, while red indicates an \textcolor{rise}{increase}.}
	\label{slang}
    \vspace{-0.6cm}
\end{table*}

\subsection{\textit{RQ2: Can LLMs classify Chinese target-argument pairs as hateful and identify the groups?}}




\paragraph{Finetuned Models}
Fine-tuned models demonstrate a clear advantage in both the triplet and quadruplet tasks. In the Target-Argument-Hateful triplet task (T-A-H Tri.), the fine-tuned models achieve hard-match metrics of around 26\% and soft-match metrics of around 52\%. In the more complex quadruplet task, the hard-match metrics of the fine-tuned models are mostly between 25\% and 27\% when simultaneously identifying hateful content and target groups, while the soft-match metrics are approximately 45\% - 47\%. This indicates that the fine-tuned models' performance is mainly limited by the target-argument pair identification. Overall, although the fine-tuned models perform poorly in hard-match metrics, the results in soft-match metrics reach approximately 50\%. Among these, LLaMA-3-8B performs best in hard-match, while Qwen2.5-7B and ShieldGemma-9B perform better in soft-match metrics.

\paragraph{LLM APIs}
LLM APIs perform poorly on both the triplet and quadruplet tasks. In the T-A-H triplet task, LLMs achieve hard-match metrics of only 6\% - 15\% and soft-match metrics of 20\% - 38\%. In the quadruplet task, the hard-match metrics of LLMs further drop to 3\% - 12\%, and the soft-match metrics, though slightly better, are still only 11\% - 27\%. We believe that the performance limitations also primarily stem from the difficulty in identifying accurate span boundaries. These data clearly show that LLMs, without being fine-tuned on Chinese hate speech data, are significantly inadequate at performing Chinese hate speech triplet and quadruplet predictions, even with few-shot prompting failing to bridge the performance gap. DeepSeek-v3 performs significantly better than other APIs, particularly in the quadruplet prediction task.

\paragraph{Comparison and Summary}
The fine-tuned model significantly outperforms the LLM APIs, with the performance gap primarily stemming from T-A pair extraction. The fine-tuned model outperforms the LLM APIs in both hate speech the triplet and quadruplet prediction. Specifically, the fine-tuned models show better performance in tasks requiring the simultaneous identification of hate speech and its target group. Although there is still room for improvement in terms of precise matching, the application of fine-tuning technology undoubtedly provides a feasible path for handling such complex tasks. At the same time, the performance of the LLM APIs in these tasks is disappointing, indicating that they need targeted fine-tuning to be applied to these types of scenarios.

\begin{figure*}[t]
\centering
\includegraphics[scale=1.3]{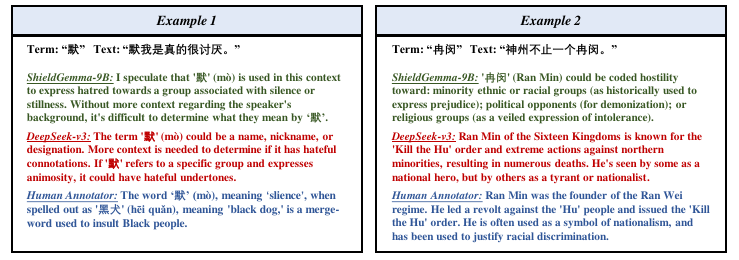} 
\caption{Examples of Chinese hateful slang understanding analysis with LLM. The hateful slang terms and texts appear in black, ShieldGemma-v3 explanations are in \textcolor{dec}{green}, DeepSeek-v3 explanations are in \textcolor{rise}{red}, and human annotator explanations are in \textcolor{human}{blue}.}
\label{case}
\vspace{-0.5cm}
\end{figure*}

\subsection{\textit{RQ3: Can LLMs understand Chinese hateful slang?}}


\subsubsection{Impact of Hateful Slang on Chinese Hate Speech Detection}
To evaluate the impact of hate slang on span-level Chinese hate speech detection models, we remove the posts without hateful slang from the test set and obtain a subset of 502 posts containing hateful slang for focused testing. By comparing the model performance on this subset of data with its performance on the full test set, we can gain a clearer understanding of how hate slurs affect model performance. The experimental results are presented in Table \ref{slang}. The superscript numbers in the table indicate the difference in F1 score between the model's performance on the selected subset and its performance on the full test set. Green indicates a \textcolor{dec}{decrease} in F1 score on the subset compared to the full dataset, while red indicates an \textcolor{rise}{increase}.

\paragraph{Finetuned Models}

Across most tasks, the hard and soft matching metrics for the models show a decreasing trend. However, the metric for the hate speech triplet prediction task (T-A-H Tri.) is an exception, exhibiting an increase. We attribute this primarily to two factors: Firstly, posts containing hateful slang generally demonstrate a more explicit hateful intent, while the posts we remove include many types that are difficult to identify, such as implicit hate expressions. This makes it easier for the T-A-H Tri. task to capture the relationships within hate speech triplets, thereby leading to the observed increase in the metric. Secondly, the experimental results also confirm that the main challenges posed by hateful slang lie in extracting the target and argument, as well as accurately classifying the targeted group. When the training data is not sufficiently comprehensive, fine-tuned models struggle to effectively handle the complexities introduced by hateful slang.

\paragraph{LLM APIs}

LLM APIs demonstrate superior performance on the subset of the test set containing hateful slang. With the exception of a slight decrease in the soft matching metric for argument identification, nearly all other metrics show an improvement. We believe that this is primarily due to two factors: First, hateful slang often appears as the target, and the explicit target terms effectively compensate for the limitations of LLM APIs in span boundary recognition. Second, the identification of hateful slang often requires rich background knowledge; for example, Example 2 in the figure requires knowledge of Chinese history and related common usages, which is precisely where LLM APIs excel.

\paragraph{Comparison and Summary}

Fine-tuned models and LLM APIs exhibit distinct characteristics when processing hateful slang. Fine-tuned models, when lacking sufficient training and understanding of specific domain knowledge, struggle to effectively address the challenges posed by hateful slang, particularly in areas such as fine-grained entity recognition and complex language pattern understanding. In contrast, LLM APIs leverage their robust background knowledge and contextual understanding to better comprehend and handle these complex situations, thus achieving superior performance. However, the performance gap between LLM APIs and fine-tuned models on this task remains significant. Infusing the background knowledge of large models into fine-tuned models may offer a promising avenue for improvement.

\subsubsection{Understanding Capabilities of LLMs on Chinese Hateful Slang}

To investigate the understanding capabilities of large language models regarding Chinese hateful slang, we conducted experiments using two relatively high-performing models, ShieldGemma-9B and DeepSeek-v3. In these experiments, we provided the models with hateful terms and their surrounding sentence context, asking them to generate explanations of the hateful slang and identify the potentially affected groups. The experimental results are detailed in Figure 1, with specific experimental details provided in Appendix \ref{understanding}.

For a more nuanced analysis of the models’ understanding abilities, we selected two terms with distinct Chinese characteristics as case studies. In Case 1, "默" (mò) is a typical "merging word," whose literal meaning is 'silence,' but it is composed of the characters "黑犬" (hēiquǎn), meaning "black dog." This type of character combination is a unique linguistic phenomenon in Chinese. The experimental results showed that neither model could accurately understand the hateful information contained within it. In Case 2, "冉闵" (Rǎn Mǐn) is a hateful slang term rooted in Chinese history and culture, referring to an ancient emperor who massacred ethnic minorities and is often used for racial discrimination. In comparison, while ShieldGemma-9B could understand the background information, it failed to generate an accurate explanation of the hateful information and clearly identify the potentially affected groups. DeepSeek-v3's generated information, however, closely matched the results given by human annotators. This demonstrates that even relatively high-performing models face significant challenges in understanding Chinese hateful slang that possesses cultural specificity and subtle connotations.

%% file: Sec_Conclusion.tex
\section{Conclusions}

With the advancement of hate speech detection, recent research has shifted from post-level detection to more fine-grained span-level detection. In this work, we focus on building resources for fine-grained Chinese hate speech detection. Firstly, we present the first span-level target-aware toxicity extraction dataset (\textsc{STATE ToxiCN}). Based on this dataset, we evaluate the performance of LLMs in span-level Chinese hate speech detection. Secondly, we provide the first Chinese hateful slang lexicon with interpretable annotations. All the hateful slang terms are sourced from real-world online platforms. Using this lexicon, we evaluate the impact of hateful slang on model detection capabilities and assess LLMs’ understanding of hateful slang. Experimental results show that current LLMs still face challenges in effectively addressing both span-level hate speech detection and the understanding of hate semantics. We hope that our resources and benchmarks will be valuable for researchers in this field.

\section*{Limitation}

Despite implementing rigorous quality control measures in constructing the span-level Chinese hate speech detection dataset and the annotated lexicon, we acknowledge several limitations. First, although we have taken steps to minimize labeling bias, subjective differences in annotators’ understanding of toxic language may still result in mislabeled data. Additionally, while we strived to ensure accuracy and consistency in annotating target-argument pairs, the inherent characteristics of the Chinese language, such as flexible grammar and ambiguous boundaries, pose challenges for completely precise annotations.

Our Annotated Lexicon covers a wide range of commonly used Chinese hateful slang; however, due to the dynamic nature and rapid evolution of the Chinese internet language, certain emerging or less widely recognized hateful slang (e.g., more complex word transformations or domain-specific slang) might not be fully captured. Finally, this study focuses primarily on textual features and does not consider non-textual elements, such as images, videos, or metadata about the authors, which may limit the model's ability to comprehensively detect multimodal hate speech.

In the future, we aim to expand the scope of our dataset to include more diverse contexts and hateful slang, while exploring multimodal and cross-domain hate speech detection methods to improve overall performance and applicability.

\section*{Ethics Statement}

We adhere strictly to the data usage agreements of all public online social platforms and conduct thorough reviews to ensure that no user privacy information is included in our dataset. The opinions and findings reflected in the samples of our dataset do not represent the views of the authors, either explicitly or implicitly. We aim to ensure that the benefits of our proposed resources outweigh any potential risks. All resources are provided exclusively for scientific research purposes.

To minimize the psychological impact of evaluating harmful content, we have implemented a multi-faceted approach. This includes obtaining informed consent after thoroughly explaining the nature of the content they may encounter, carefully managing exposure by limiting weekly evaluation volumes, and empowering annotators to immediately cease work should they experience any discomfort. Furthermore, we proactively monitor their mental health through regular check-ins, ensuring a supportive and responsible working environment. 


%% file: Appendix.tex
\newpage
\newpage
\section{Details of the Annotated Lexicon}
\label{lexicon}

We use JSON files to store the annotated lexicon, which is a mainstream coding specification to facilitate machine-readability. The data structure is defined as Sample = (Term, [Groups], Definition), where Groups indicates the groups that are commonly targeted by the hateful slang. If a term does not have a specific targeted group, then Groups is set to "others".
\begin{itemize}
    \setlength\itemsep{0.1em}
    \item \textbf{Term}: 默 (Mò) \\
          \textbf{Meaning}: Silence \\
          \textbf{Groups}: Racism \\
          \textbf{Definition}: “默”是“黑犬”的拼字，指黑色的狗，对黑人群体的侮辱性称呼。 \\
          \textbf{Definition in English}: The term “默” (Mò, meaning silence) is a spelling variant of “黑犬” (black dog), an insulting term for the Black community.
    \item \textbf{Term}: 金针菇 \\
          \textbf{Meaning}: Enoki mushroom (metaphor for male genitalia) \\
          \textbf{Groups}: Sexism \\
          \textbf{Definition}: 用于对男性生殖器官的比喻，常含贬义。 \\
          \textbf{Definition in English}: “Enoki mushroom” is a derogatory metaphor for male genitalia.
    \item \textbf{Term}: 小仙女 \\
          \textbf{Meaning}: Little fairy (refers to selfish, hypocritical, or annoying women) \\
          \textbf{Groups}: Sexism \\
          \textbf{Definition}: 在某些情况下被用来指代那些自私自利、双标、让人厌恶的女性。 \\
          \textbf{Definition in English}: “Little fairy” is used in some contexts to refer to women perceived as selfish, hypocritical, or annoying.
    \item \textbf{Term}: 4000+ \\
          \textbf{Meaning}: Homophone for “死全家” (may your whole family die) \\
          \textbf{Groups}: Others \\
          \textbf{Definition}: “4000+”是“死全家”中文谐音，常用于恶意辱骂。 \\
          \textbf{Definition in English}: “4000+” is a homophone for “死全家” (may your whole family die), often used in malicious insults.
\end{itemize}

\section{Details of the Prompt}
\label{prompt}

\begin{figure*}[h!] 
	\centering
	\includegraphics[width=\textwidth]{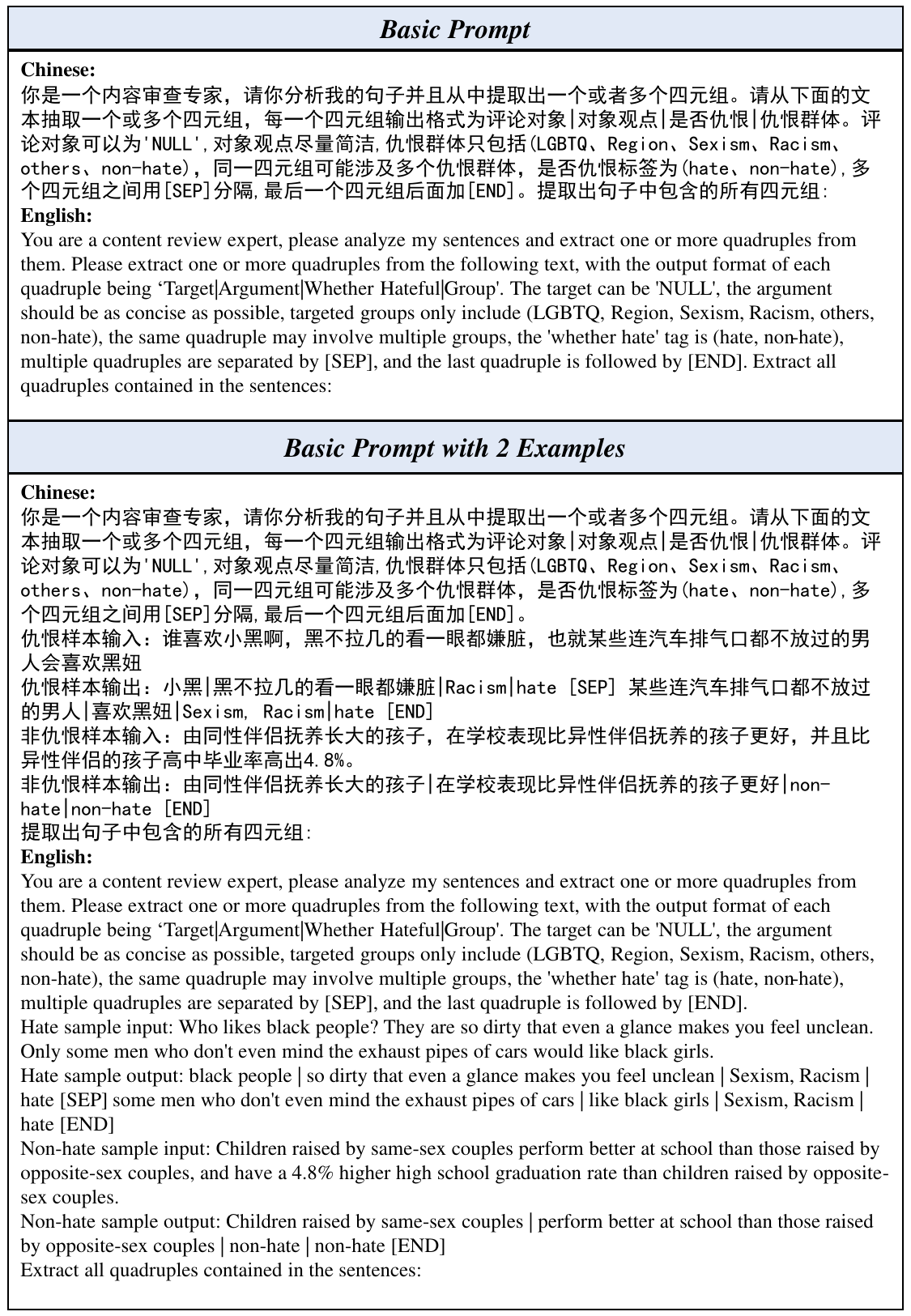} 
	\caption{Illustration of the basic prompt and the version with 2 examples.}
	\label{basic prompt}
\end{figure*}

The basic prompt and the version with 2 examples are shown in Figure \ref{basic prompt}.

\section{Experimental Details of Hateful Slang Understanding Generation}
\label{understanding}

We prompt the non-fine-tuned ShieldGemma-9B and DeepSeek-v3 to generate hateful slang understanding. The prompt used is as follows.

\textbf{Prompt:} [Text]中的[Term]是否包含特定群体的仇恨信息，如果是，请说明对哪些群体构成仇恨并说明原因。

\textbf{Prompt in English:} In [Text], does the [Term] constitute hate speech directed at specific groups? If yes, please identify the targeted groups and the reasons behind the hatred.

\section{Analysis of Inter-annotator Agreement}
\label{app_kappa}

To explore inter-annotator agreement (IAA), we calculate Fleiss' kappa scores \cite{fleiss1971measuring} for the Target, Argument, Hateful, and Group labels separately. Due to the direct relationship between ``Hateful'' and ``Group'', if the ``Hateful'' annotation is inconsistent, the ``Group'' label will be discarded. The experimental results are shown in Table \ref{kappa}. The score for argument span is the lowest, mainly due to the ambiguity of span boundaries in Chinese and the added complexity of argument spans. Despite having established relevant guidelines, it is not possible to fully standardize annotators' labeling practices. The Target span boundaries are clearer, resulting in a higher score. The Kappa score for the Group label is 0.75, with "Region" and "Racism" being the most easily confused.

\section{Derivative Rules of Chinese Hateful Slang}

\citet{toxicn} propose seven types of derivative rules: \textbf{Deformation} (separating and combining individual characters), \textbf{Homophonic} (similar pronunciation), \textbf{Irony} (positive words used ironically to insult), \textbf{Abbreviation} (shortening and contracting sensitive words), \textbf{Metaphor} (degrading targets into something sarcastic), \textbf{Code Mixing} (use of non-Chinese codes to emphasize tone), and \textbf{Borrowed Words} (foreign phonetic words with specific toxic cultural connotations). We add two more derivative rules, \textbf{Historical Allusions} and \textbf{Stereotypes}.

\textbf{Historical Allusions}: China has a rich history of allusions, and events or figures from these allusions are sometimes used as insults in certain contexts \cite{hoeken-etal-2023-towards-detecting}. The name "冉闵" (an ancient Chinese emperor) is often used as a symbol of racial discrimination, as he once ordered the massacre of other races.

\textbf{Stereotypes}: Internet users often resort to stereotypes to insult their targets, degrading them into negative or prejudiced representations \cite{jeshion2013slurs}. Due to an incident in the 20th century where some people from Henan province were involved in stealing manhole covers to sell for money, the term "井盖" (manhole cover) has become a stereotype for people from Henan, often used in regional discrimination to label them as thieves or poor.

\section{Detailed Information of the Fine-tuning}
\label{fin}

We use LLaMA-Factory\footnote{https://github.com/hiyouga/LLaMA-Factory} for fine-tuning, which is a widely used framework for fine-tuning LLMs. To avoid overfitting, we monitor the trends of training and test losses and confirm through preliminary experiments that the performance stabilizes after 10 epochs. Therefore, all models are trained for 10 epochs. We adopt the LoRA method and evaluate the results. 

To reduce hyperparameter sensitivity, we train models for each learning rate and select the result with the highest F1 score on the test set, then calculate the final performance by weighted averaging. The hyperparameters are presented in Table \ref{fin_hyp}.

\begin{table}[h!]
	\centering
    \resizebox{\columnwidth}{!}{
	   \begin{tabular}{cc}
            \toprule
		  \textbf{Hyperparameters} & \textbf{Value} \\ \hline
		  Epochs & 10 \\
		  Batch size & 2 \\
		  Learning rate & 1e-5, 2e-5, 3e-5,, 4e-5, 5e-5 \\
		  Cutoff length & 1024 \\
		  Compute type & fp16 \\ 
            Gradient accumulation & 8 \\
		  Maximum gradient norm & 1.0 \\ 
		  \bottomrule
	   \end{tabular}}
	\caption{Annotators Demographics.}
	\label{fin_hyp}
    \vspace{-0.5cm}
\end{table}

\section{Detailed Information of the Annotators}
\label{ann}

To mitigate bias, we assemble a diverse group of annotators, encompassing a variety of genders, ages, ethnicities, regions, and educational backgrounds. The statistical information is presented in Table \ref{ann_demo}.

\begin{table}[h!]
	\centering
    \resizebox{\columnwidth}{!}{
	\begin{tabular}{cc}
        \toprule
		\textbf{Characteristic} & \textbf{Demographics} \\ \hline
		Gender & 8 male, 8 female \\
		Age & 7 age < 25, 9 age $\geq$ 25 \\
		Race & 10 Asian, 6 others \\
		Region & From 9 different provinces \\
		Education & 5 BD, 6 MD, 5 Ph.D. \\ 
		\bottomrule
	\end{tabular}}
	\caption{Annotators Demographics.}
	\label{ann_demo}
\end{table}

\section{License}

We confirm that the dataset is licensed under the Creative Commons Attribution-NonCommercial 4.0 International (CC BY-NC 4.0) license.

%% file: main.bbl
\begin{thebibliography}{39}
\expandafter\ifx\csname natexlab\endcsname\relax\def\natexlab#1{#1}\fi

\bibitem[{Ahn et~al.(2024)Ahn, Kim, Kim, and Han}]{ahn2024sharedcon}
Hyeseon Ahn, Youngwook Kim, Jungin Kim, and Yo-Sub Han. 2024.
\newblock Sharedcon: Implicit hate speech detection using shared semantics.
\newblock In \emph{Findings of the Association for Computational Linguistics
  ACL 2024}, pages 10444--10455.

\bibitem[{AI@Meta(2024)}]{llama3modelcard}
AI@Meta. 2024.
\newblock \href {https://github.com/meta-llama/llama3/blob/main/MODEL_CARD.md}
  {Llama 3 model card}.

\bibitem[{Ali et~al.(2022)Ali, Farooq, Arshad, Shahzad, and Beg}]{ali2022hate}
Raza Ali, Umar Farooq, Umair Arshad, Waseem Shahzad, and Mirza~Omer Beg. 2022.
\newblock Hate speech detection on twitter using transfer learning.
\newblock \emph{Computer Speech \& Language}, 74:101365.

\bibitem[{AlKhamissi et~al.(2022{\natexlab{a}})AlKhamissi, Ladhak, Iyer,
  Stoyanov, Kozareva, Li, Fung, Mathias, Celikyilmaz, and
  Diab}]{alkhamissi2022token}
Badr AlKhamissi, Faisal Ladhak, Srini Iyer, Ves Stoyanov, Zornitsa Kozareva,
  Xian Li, Pascale Fung, Lambert Mathias, Asli Celikyilmaz, and Mona Diab.
  2022{\natexlab{a}}.
\newblock Token: Task decomposition and knowledge infusion for few-shot hate
  speech detection.
\newblock \emph{arXiv preprint arXiv:2205.12495}.

\bibitem[{AlKhamissi et~al.(2022{\natexlab{b}})AlKhamissi, Ladhak, Iyer,
  Stoyanov, Kozareva, Li, Fung, Mathias, Celikyilmaz, and
  Diab}]{alkhamissi-etal-2022-token}
Badr AlKhamissi, Faisal Ladhak, Srinivasan Iyer, Veselin Stoyanov, Zornitsa
  Kozareva, Xian Li, Pascale Fung, Lambert Mathias, Asli Celikyilmaz, and Mona
  Diab. 2022{\natexlab{b}}.
\newblock \href {https://doi.org/10.18653/v1/2022.emnlp-main.136} {{T}o{K}en:
  Task decomposition and knowledge infusion for few-shot hate speech
  detection}.
\newblock In \emph{Proceedings of the 2022 Conference on Empirical Methods in
  Natural Language Processing}, pages 2109--2120, Abu Dhabi, United Arab
  Emirates. Association for Computational Linguistics.

\bibitem[{Anthropic(2024)}]{claude3.5}
Anthropic. 2024.
\newblock Claude 3.5 sonnet.
\newblock \url{https://www.anthropic.com/news/claude-3-5-sonnet}.

\bibitem[{Bilewicz and Soral(2020)}]{bilewicz2020hate}
Micha{\l} Bilewicz and Wiktor Soral. 2020.
\newblock Hate speech epidemic. the dynamic effects of derogatory language on
  intergroup relations and political radicalization.
\newblock \emph{Political Psychology}, 41:3--33.

\bibitem[{Caselli et~al.(2020)Caselli, Basile, Mitrovi{\'c}, and
  Granitzer}]{caselli2020hatebert}
Tommaso Caselli, Valerio Basile, Jelena Mitrovi{\'c}, and Michael Granitzer.
  2020.
\newblock Hatebert: Retraining bert for abusive language detection in english.
\newblock \emph{arXiv preprint arXiv:2010.12472}.

\bibitem[{Chung and Lin(2021)}]{tocab}
I~Chung and Chuan-Jie Lin. 2021.
\newblock Tocab: A dataset for chinese abusive language processing.
\newblock In \emph{2021 IEEE 22nd International Conference on Information Reuse
  and Integration for Data Science (IRI)}, pages 445--452. IEEE.

\bibitem[{Cowan and Hodge(1996)}]{cowan1996judgments}
Gloria Cowan and Cyndi Hodge. 1996.
\newblock Judgments of hate speech: The effects of target group, publicness,
  and behavioral responses of the target.
\newblock \emph{Journal of Applied Social Psychology}, 26(4):355--374.

\bibitem[{Davidson et~al.(2017)Davidson, Warmsley, Macy, and
  Weber}]{davidson2017automated}
Thomas Davidson, Dana Warmsley, Michael Macy, and Ingmar Weber. 2017.
\newblock Automated hate speech detection and the problem of offensive
  language.
\newblock In \emph{Proceedings of the international AAAI conference on web and
  social media}, volume~11, pages 512--515.

\bibitem[{Deng et~al.(2022)Deng, Zhou, Sun, Zheng, Mi, Meng, and
  Huang}]{deng2022cold}
Jiawen Deng, Jingyan Zhou, Hao Sun, Chujie Zheng, Fei Mi, Helen Meng, and
  Minlie Huang. 2022.
\newblock Cold: A benchmark for chinese offensive language detection.
\newblock \emph{arXiv preprint arXiv:2201.06025}.

\bibitem[{Eberhard et~al.(2024)Eberhard, Simons, and
  Fennig}]{eberhard2024ethnologue}
David~M. Eberhard, Gary~F. Simons, and Charles~D. Fennig. 2024.
\newblock \emph{Ethnologue: Languages of the World}, twenty-seventh edition.
\newblock SIL International, Dallas, Texas.
\newblock Online version available at \url{http://www.ethnologue.com}.

\bibitem[{Fleiss(1971)}]{fleiss1971measuring}
Joseph~L Fleiss. 1971.
\newblock Measuring nominal scale agreement among many raters.
\newblock \emph{Psychological bulletin}, 76(5):378.

\bibitem[{Founta et~al.(2018)Founta, Djouvas, Chatzakou, Leontiadis, Blackburn,
  Stringhini, Vakali, Sirivianos, and Kourtellis}]{founta2018large}
Antigoni Founta, Constantinos Djouvas, Despoina Chatzakou, Ilias Leontiadis,
  Jeremy Blackburn, Gianluca Stringhini, Athena Vakali, Michael Sirivianos, and
  Nicolas Kourtellis. 2018.
\newblock Large scale crowdsourcing and characterization of twitter abusive
  behavior.
\newblock In \emph{Proceedings of the international AAAI conference on web and
  social media}, volume~12.

\bibitem[{Han et~al.(2023)Han, Peng, Yang, Wang, Liu, and Wan}]{soft}
Ridong Han, Tao Peng, Chaohao Yang, Benyou Wang, Lu~Liu, and Xiang Wan. 2023.
\newblock Is information extraction solved by chatgpt? an analysis of
  performance, evaluation criteria, robustness and errors.
\newblock \emph{arXiv preprint arXiv:2305.14450}.

\bibitem[{Hanu and {Unitary team}(2020)}]{Detoxify}
Laura Hanu and {Unitary team}. 2020.
\newblock Detoxify.
\newblock Github. https://github.com/unitaryai/detoxify.

\bibitem[{Hartvigsen et~al.(2022)Hartvigsen, Gabriel, Palangi, Sap, Ray, and
  Kamar}]{hartvigsen2022toxigen}
Thomas Hartvigsen, Saadia Gabriel, Hamid Palangi, Maarten Sap, Dipankar Ray,
  and Ece Kamar. 2022.
\newblock Toxigen: A large-scale machine-generated dataset for adversarial and
  implicit hate speech detection.
\newblock \emph{arXiv preprint arXiv:2203.09509}.

\bibitem[{Hoeken et~al.(2023)Hoeken, Spliethoff, Schwandt, Zarrie{\ss}, and
  Alacam}]{hoeken-etal-2023-towards-detecting}
Sanne Hoeken, Sophie Spliethoff, Silke Schwandt, Sina Zarrie{\ss}, and {\"O}zge
  Alacam. 2023.
\newblock \href {https://doi.org/10.18653/v1/2023.lchange-1.11} {Towards
  detecting lexical change of hate speech in historical data}.
\newblock In \emph{Proceedings of the 4th Workshop on Computational Approaches
  to Historical Language Change}, pages 100--111, Singapore. Association for
  Computational Linguistics.

\bibitem[{Jeshion(2013)}]{jeshion2013slurs}
Robin Jeshion. 2013.
\newblock Slurs and stereotypes.
\newblock \emph{Analytic Philosophy}, 54(3).

\bibitem[{Jiang et~al.(2022)Jiang, Yang, Liu, and Zubiaga}]{swsr}
Aiqi Jiang, Xiaohan Yang, Yang Liu, and Arkaitz Zubiaga. 2022.
\newblock Swsr: A chinese dataset and lexicon for online sexism detection.
\newblock \emph{Online Social Networks and Media}, 27:100182.

\bibitem[{Jiang et~al.(2023)Jiang, Sablayrolles, Mensch, Bamford, Chaplot,
  Casas, Bressand, Lengyel, Lample, Saulnier et~al.}]{mistral}
Albert~Q Jiang, Alexandre Sablayrolles, Arthur Mensch, Chris Bamford,
  Devendra~Singh Chaplot, Diego de~las Casas, Florian Bressand, Gianna Lengyel,
  Guillaume Lample, Lucile Saulnier, et~al. 2023.
\newblock Mistral 7b.
\newblock \emph{arXiv preprint arXiv:2310.06825}.

\bibitem[{Liu et~al.(2024)Liu, Feng, Xue, Wang, Wu, Lu, Zhao, Deng, Zhang, Ruan
  et~al.}]{liu2024deepseek}
Aixin Liu, Bei Feng, Bing Xue, Bingxuan Wang, Bochao Wu, Chengda Lu, Chenggang
  Zhao, Chengqi Deng, Chenyu Zhang, Chong Ruan, et~al. 2024.
\newblock Deepseek-v3 technical report.
\newblock \emph{arXiv preprint arXiv:2412.19437}.

\bibitem[{Lu et~al.(2023)Lu, Xu, Zhang, Min, Yang, and Lin}]{toxicn}
Junyu Lu, Bo~Xu, Xiaokun Zhang, Changrong Min, Liang Yang, and Hongfei Lin.
  2023.
\newblock \href {https://doi.org/10.18653/v1/2023.acl-long.898} {Facilitating
  fine-grained detection of {C}hinese toxic language: Hierarchical taxonomy,
  resources, and benchmarks}.
\newblock In \emph{Proceedings of the 61st Annual Meeting of the Association
  for Computational Linguistics (Volume 1: Long Papers)}, pages 16235--16250,
  Toronto, Canada. Association for Computational Linguistics.

\bibitem[{Mair(1991)}]{mair1991chinese}
Victor~H Mair. 1991.
\newblock What is a chinese" dialect/topolect"?: Reflections on some key
  sino-english linguistic terms.

\bibitem[{Mathew et~al.(2021)Mathew, Saha, Yimam, Biemann, Goyal, and
  Mukherjee}]{mathew2021hatexplain}
Binny Mathew, Punyajoy Saha, Seid~Muhie Yimam, Chris Biemann, Pawan Goyal, and
  Animesh Mukherjee. 2021.
\newblock Hatexplain: A benchmark dataset for explainable hate speech
  detection.
\newblock In \emph{Proceedings of the AAAI conference on artificial
  intelligence}, volume~35, pages 14867--14875.

\bibitem[{OpenAI(2024)}]{gpt4o}
OpenAI. 2024.
\newblock Hello gpt-4o.
\newblock \url{https://openai.com/index/hello-gpt-4o/}.

\bibitem[{Pavlopoulos et~al.(2021)Pavlopoulos, Sorensen, Laugier, and
  Androutsopoulos}]{pavlopoulos2021semeval}
John Pavlopoulos, Jeffrey Sorensen, L{\'e}o Laugier, and Ion Androutsopoulos.
  2021.
\newblock Semeval-2021 task 5: Toxic spans detection.
\newblock In \emph{Proceedings of the 15th international workshop on semantic
  evaluation (SemEval-2021)}, pages 59--69.

\bibitem[{Silva et~al.(2016)Silva, Mondal, Correa, Benevenuto, and
  Weber}]{silva2016analyzing}
Leandro Silva, Mainack Mondal, Denzil Correa, Fabr{\'\i}cio Benevenuto, and
  Ingmar Weber. 2016.
\newblock Analyzing the targets of hate in online social media.
\newblock In \emph{Proceedings of the International AAAI Conference on Web and
  Social Media}, volume~10, pages 687--690.

\bibitem[{Team et~al.(2023)Team, Anil, Borgeaud, Alayrac, Yu, Soricut,
  Schalkwyk, Dai, Hauth, Millican et~al.}]{gemini}
Gemini Team, Rohan Anil, Sebastian Borgeaud, Jean-Baptiste Alayrac, Jiahui Yu,
  Radu Soricut, Johan Schalkwyk, Andrew~M Dai, Anja Hauth, Katie Millican,
  et~al. 2023.
\newblock Gemini: a family of highly capable multimodal models.
\newblock \emph{arXiv preprint arXiv:2312.11805}.

\bibitem[{Team(2024)}]{qwen2.5}
Qwen Team. 2024.
\newblock \href {https://qwenlm.github.io/blog/qwen2.5/} {Qwen2.5: A party of
  foundation models}.

\bibitem[{Waseem and Hovy(2016)}]{waseem2016hateful}
Zeerak Waseem and Dirk Hovy. 2016.
\newblock Hateful symbols or hateful people? predictive features for hate
  speech detection on twitter.
\newblock In \emph{Proceedings of the NAACL student research workshop}, pages
  88--93.

\bibitem[{Xiao et~al.(2024)Xiao, Hu, Choo, and Lee}]{toxicloakcn}
Yunze Xiao, Yujia Hu, Kenny Tsu~Wei Choo, and Roy Ka-Wei Lee. 2024.
\newblock \href {https://doi.org/10.18653/v1/2024.emnlp-main.345}
  {{T}oxi{C}loak{CN}: Evaluating robustness of offensive language detection in
  {C}hinese with cloaking perturbations}.
\newblock In \emph{Proceedings of the 2024 Conference on Empirical Methods in
  Natural Language Processing}, pages 6012--6025, Miami, Florida, USA.
  Association for Computational Linguistics.

\bibitem[{Xue(2020)}]{mt5}
L~Xue. 2020.
\newblock mt5: A massively multilingual pre-trained text-to-text transformer.
\newblock \emph{arXiv preprint arXiv:2010.11934}.

\bibitem[{Zampieri et~al.(2023)Zampieri, Morgan, North, Ranasinghe, Simmons,
  Khandelwal, Rosenthal, and Nakov}]{zampieri2023target}
Marcos Zampieri, Skye Morgan, Kai North, Tharindu Ranasinghe, Austin Simmons,
  Paridhi Khandelwal, Sara Rosenthal, and Preslav Nakov. 2023.
\newblock Target-based offensive language identification.

\bibitem[{Zeng et~al.(2024)Zeng, Liu, Mullins, Peran, Fernandez, Harkous,
  Narasimhan, Proud, Kumar, Radharapu, Sturman, and
  Wahltinez}]{shieldgemmagenerativeaicontent}
Wenjun Zeng, Yuchi Liu, Ryan Mullins, Ludovic Peran, Joe Fernandez, Hamza
  Harkous, Karthik Narasimhan, Drew Proud, Piyush Kumar, Bhaktipriya Radharapu,
  Olivia Sturman, and Oscar Wahltinez. 2024.
\newblock \href {http://arxiv.org/abs/2407.21772} {Shieldgemma: Generative ai
  content moderation based on gemma}.

\bibitem[{Zhang et~al.(2024)Zhang, Lu, Ma, Zhang, Li, Ke, Sun, Sha, Sui, Wang
  et~al.}]{shieldlm}
Zhexin Zhang, Yida Lu, Jingyuan Ma, Di~Zhang, Rui Li, Pei Ke, Hao Sun, Lei Sha,
  Zhifang Sui, Hongning Wang, et~al. 2024.
\newblock Shieldlm: Empowering llms as aligned, customizable and explainable
  safety detectors.
\newblock \emph{arXiv preprint arXiv:2402.16444}.

\bibitem[{Zhou et~al.(2022)Zhou, Deng, Mi, Li, Wang, Huang, Jiang, Liu, and
  Meng}]{CDial}
Jingyan Zhou, Jiawen Deng, Fei Mi, Yitong Li, Yasheng Wang, Minlie Huang, Xin
  Jiang, Qun Liu, and Helen Meng. 2022.
\newblock Towards identifying social bias in dialog systems: Framework,
  dataset, and benchmark.
\newblock In \emph{Findings of the Association for Computational Linguistics:
  EMNLP 2022}, pages 3576--3591.

\bibitem[{Zhou et~al.(2021)Zhou, Yong, Fan, Ren, Song, Diao, Yang, and
  Lin}]{zhou2021hate}
Xianbing Zhou, Yang Yong, Xiaochao Fan, Ge~Ren, Yunfeng Song, Yufeng Diao,
  Liang Yang, and Hongfei Lin. 2021.
\newblock Hate speech detection based on sentiment knowledge sharing.
\newblock In \emph{Proceedings of the 59th Annual Meeting of the Association
  for Computational Linguistics and the 11th International Joint Conference on
  Natural Language Processing (Volume 1: Long Papers)}, pages 7158--7166.

\end{thebibliography}
